\pdfoutput=1

\documentclass[11pt]{article}

\usepackage{ACL2023}

\usepackage{times}
\usepackage{latexsym}

\usepackage[T1]{fontenc}

\usepackage[utf8]{inputenc}

\usepackage{microtype}

\usepackage{inconsolata}

\usepackage{graphicx} 
\usepackage{booktabs}
\usepackage{tabularx} 
\usepackage{longtable}

\usepackage{tcolorbox}

\title{Learning Wisdom from Errors: Promoting LLM's Continual Relation Learning through Exploiting Error Cases}

\author{
Shaozhe Yin$^{1}$,
Jinyu Guo$^{2}$,
Kai Shuang$^{1}$,
Xia Liu$^{3}$,
Ruize Ou$^{1}$,
 \\
$^{1}$Beijing University of Posts and Telecommunications, Beijing, China \\
$^{2}$University of Electronic Science and Technology of China, Chengdu, China \\
$^{3}$China National Institute of Standardization, Beijing, China \\
$^{1}$\texttt{\{yinshaozhe, shuangk, orzzz\}@bupt.edu.cn}, $^{2}$\texttt{guojinyu@uestc.edu.cn} \\
$^{3}$\texttt{liuxia1010@163.com}
}

\begin{document}
\maketitle
\begin{abstract}
Continual Relation Extraction (CRE) aims to continually learn new emerging relations while avoiding catastrophic forgetting. Existing CRE methods mainly use memory replay and contrastive learning to mitigate catastrophic forgetting. However, these methods do not attach importance to the error cases that can reveal the model's cognitive biases more effectively. To address this issue, we propose an instruction-based continual contrastive tuning approach for Large Language Models (LLMs) in CRE. Different from existing CRE methods that typically handle the training and memory data in a unified manner, this approach splits the training and memory data of each task into two parts respectively based on the correctness of the initial responses and treats them differently through dual-task fine-tuning. In addition, leveraging the advantages of LLM's instruction-following ability, we propose a novel instruction-based contrastive tuning strategy for LLM to continuously correct current cognitive biases with the guidance of previous data in an instruction-tuning manner, which mitigates the gap between old and new relations in a more suitable way for LLMs. We experimentally evaluate our model on TACRED and FewRel, and the results show that our model achieves new state-of-the-art CRE performance with significant improvements, demonstrating the importance of specializing in exploiting error cases.
\end{abstract}

\section{Introduction}

The Relation Extraction (RE) task aims to identify the relation between two given entities in a text. Conventional relation extraction models learn all the data simultaneously and make predictions. Nevertheless, in real-world scenarios, new relations continuously emerge, which requires models to continuously adapt to these new relations while preserving performance on previously learned ones. This demand for ongoing adaptability has led to the development of Continual Relation Extraction (CRE). Despite the advancements in methods within this field, CRE models often suffer from catastrophic forgetting, where performance on previously learned relations declines significantly after learning new ones.

\begin{figure}
    \centering
    \includegraphics[width=1\linewidth]{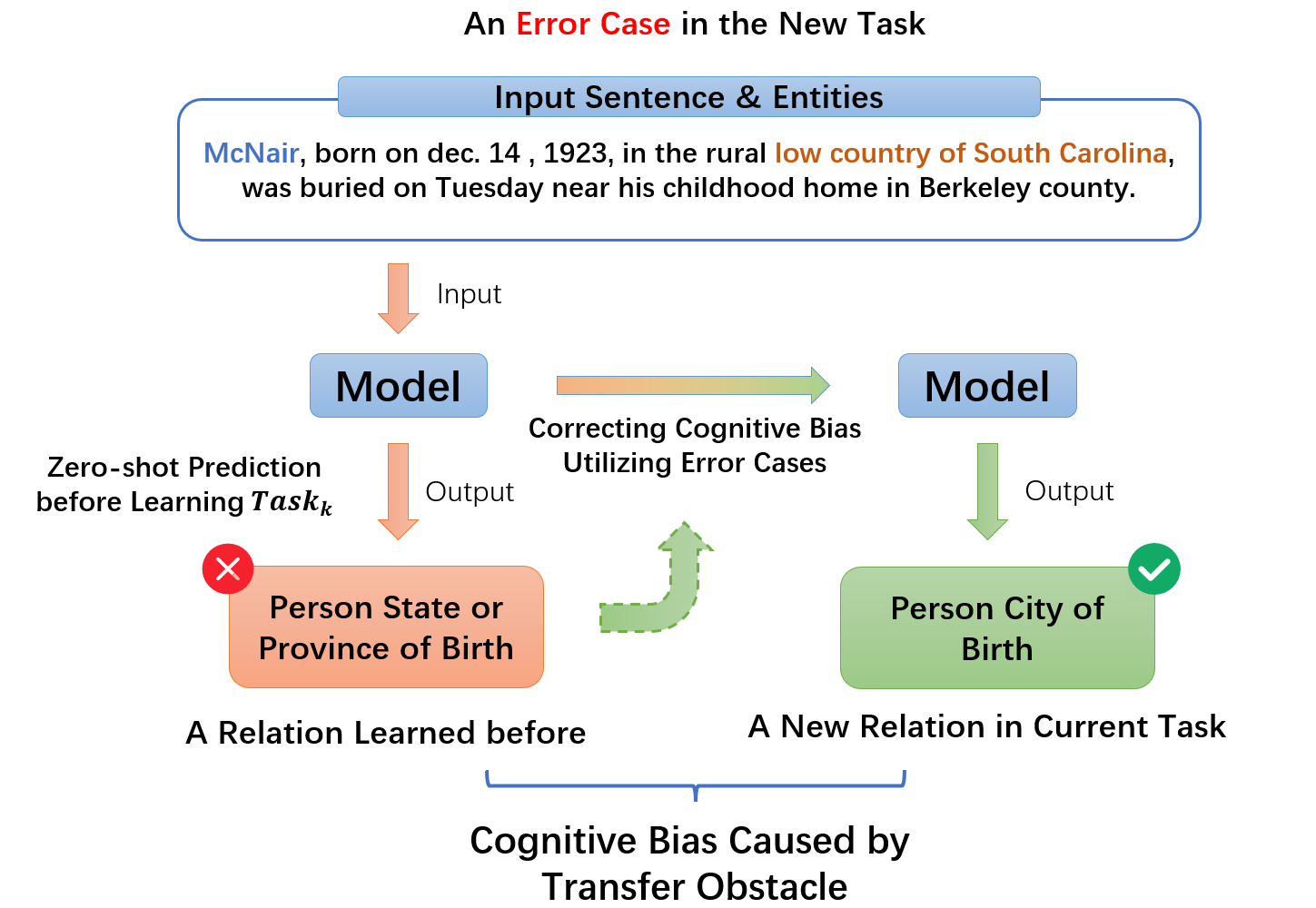}
    \caption{The illustration of an error case caused by cognitive bias in CRE. We make the model predict the sample in the new task before learning it. The error cases are used to correct current cognitive biases.}
    \label{fig:intro}
\end{figure}

In order to alleviate catastrophic forgetting, some recent research have delved into the reasons behind it. For instance, some research consider that it lies in the confusion between analogous relations \cite{Zhao2022ConsistentRL, Hu2022ImprovingCR, Zhao2023ImprovingCR_r:12, Xiong2023RationaleEnhancedLM}. However, regardless of the reason, the characteristics of staged learning make it difficult for the model to fully adapt to the distribution of all data, resulting in overfitting to a specific subset. We refer to this phenomenon as the cognitive bias among different subsets. 

Obviously, this cognitive bias can be well reflected by the error cases generated by the model during learning. Therefore, exploiting error cases as a key to identify and correct cognitive biases can help the model transfer knowledge more efficiently between tasks, thereby alleviating catastrophic forgetting, as illustrated in Figure \ref{fig:intro}. Error cases enable us to adjust training strategies dynamically and help the model overcome the transfer obstacle. Despite the aforementioned importance of error cases, previous CRE research did not focus on the unique value of error cases, resulting in their potential in model optimization not being fully explored.

On the other hand, some research have demonstrated that large language models (LLMs), owing to their robust language comprehension and instruction-following capabilities, can learn from their mistakes and enhance their performance \cite{Madaan2023SelfRefineIR_r:17, yang-etal-2023-failures, Zhang2024InContextPL, C-ICL, tong-etal-2024-llms}. Given this characteristic, we believe that LLM's instruction-following ability enables them to continuously analyze their own cognitive biases in error cases and then self-correct them in a more suitable way.

Based on the above discussion, we propose a continual \textbf{C}ontrastive \textbf{I}nstruction \textbf{T}uning method for CRE (\textbf{CIT-CRE}) based on LLM, which enables the model to continuously correct the cognitive biases by exploiting error cases. Specifically, we first divide the training data into two sets based on whether the model can directly inference and answer the training data of the new task correctly. After this construction, we leverage differentiated mechanisms for easy sets and hard sets to optimize the model. For easy samples that the model can directly answer correctly, we use a simple instruction to maintain its correct understanding, as the model has already mastered this knowledge. For hard samples where the model answers incorrectly, we devise a novel contrastive instruction tuning strategy by utilizing both the previous easy and hard memory data in different manners: 1) first, we facilitate analogical learning from prior easy memory data. Concretely, we retrieve semantically similar samples from the easy memory data associated with the most similar relation and then guide the model to apply and transfer previously learned knowledge by adding the corresponding similar relation-analytical prompts; 2) second, to enable the model to learn wisdom from past mistakes, we retrieve samples from the hard memory data based on the similarity of error reasons identified by an external LLM and guide the model to avoid these similar mistakes by adding corresponding guidance prompts. Finally, we sample from two training data sets to update two corresponding memory sets for subsequent tasks. Unlike previous research on contrastive learning that mainly focus on loss functions, we implement the idea of contrast through instructions, which aligns more closely with the ability of instruction-following of LLMs \cite{instruction-tuning}. Through the process of jointly using two memory data as described above, we enable the model to correct cognitive biases, thereby connecting new and old tasks and mitigating catastrophic forgetting.

We evaluate our model on TACRED and FewRel, and the results show that our model significantly outperforms the state-of-the-art CRE models. In particular, due to continuous correction for cognitive bias, our model demonstrates increasing advantages as the number of learning tasks increases.

In summary, our contributions are as follows:

\begin{itemize}
\item We split the training data and memory data into two parts for differentiated processing. Then we focus on the error cases and utilize two types of memory data for joint training on them to correct the model's cognitive bias.
\item 
Different from previous research on contrastive learning that mainly focus on loss functions, we construct a contrastive instruction tuning strategy for hard samples, which leverages LLM's instruction-following abilities to continuously correct cognitive biases more effectively.
\item We conduct experiments on two mainstream CRE datasets, TACRED and FewRel, and our method achieves state-of-the-art performance on CRE with significant improvements.
\end{itemize}

\section{Related Work}

Continual learning enables models to sequentially learn from data while adapting to changing distributions \cite{wang2024comprehensivesurveycontinuallearning}. Current methods fall into three categories: (1) Regularization-based methods, which constrain parameter updates to retain prior knowledge \cite{Learningwithoutforgetting,r:3}; (2) Dynamic architecture methods, which expand the model to accommodate new tasks without forgetting old ones \cite{r:4,r:5}; and (3) Memory-based methods, which replay data from old tasks to mitigate forgetting \cite{r:7,r:8}. In continual relation extraction (CRE), memory-based methods are dominant due to their superior performance \cite{Wang2019SentenceEA_r:22,Han2020ContinualRL_r:23,Wu2021CurriculumMetaLF_r:24,Cui2021RefiningSE_r:25,Zhao2022ConsistentRL,Hu2022ImprovingCR,Zhao2023ImprovingCR_r:12, huang-etal-2024-dp}. Recent studies highlight that catastrophic forgetting in CRE often stems from confusion between analogous relations \cite{Zhao2022ConsistentRL,Hu2022ImprovingCR,Zhao2023ImprovingCR_r:12,Xiong2023RationaleEnhancedLM}. To address this, \cite{Zhao2022ConsistentRL,Hu2022ImprovingCR} use contrastive learning to maintain relation distinguishability, while \cite{Xiong2023RationaleEnhancedLM} employ contrastive replay for analogous relation separation.

Meanwhile, research on LLMs has revealed that scaling model parameters and employing instruction-tuning can unlock emergent capabilities, significantly enhancing language understanding \cite{instruction-tuning,Zhao2023ASO_r:16}. Recent work explores leveraging LLMs for self-correction (reflection). For instance, \cite{Madaan2023SelfRefineIR_r:17} uses feedback from identical models to iteratively refine outputs, while \cite{wang-li-2023-learning,C-ICL,Zhang2024InContextPL, tong-etal-2024-llms} investigate how models learn from their own errors. Additionally, \cite{Pang2023GuidelineLF, yang-etal-2023-failures} guides model predictions through guidelines derived from reflecting on incorrect cases.

Based on the above two aspects, we implement the idea of contrast through instructions, enabling the LLM to understand the features of error cases more comprehensively and thus correct cognitive biases more effectively.

\section{Method}
\subsection{Task Formulation}
The CRE task involves a sequence of tasks, defined as $T = \left\{ T_{1}, T_{2},\ldots, T_{k} \right\}$. Each individual task is a conventional relation extraction task, with its corresponding relation set and dataset denoted as $R_k$ and $D_{k} = \left\{ \left( x_{i},r_{i} \right) \right\}_{i=1}^N$, where $x_i$ includes the sentence text and its corresponding two entities, and $r_i \in R_k$ is the relation corresponding to $x_i$. A continual relation extraction model needs to perform well on all seen tasks $\tilde{T_{k}} = {\bigcup_{i = 1}^{k}T_{i}}$  where the set of all seen relations is defined as $\tilde{R_{k}} = {\bigcup_{i = 1}^{k}R_{i}}$. For each relation $r$, we select a fixed-size memory data $M_r$. The memory data for all learned relations is represented as $\tilde{M_{k}} = {\bigcup_{r \in \tilde{R_{k}}}M_{r}}$.

\begin{figure*}[t]
    \centering
    \includegraphics[width=1\linewidth]{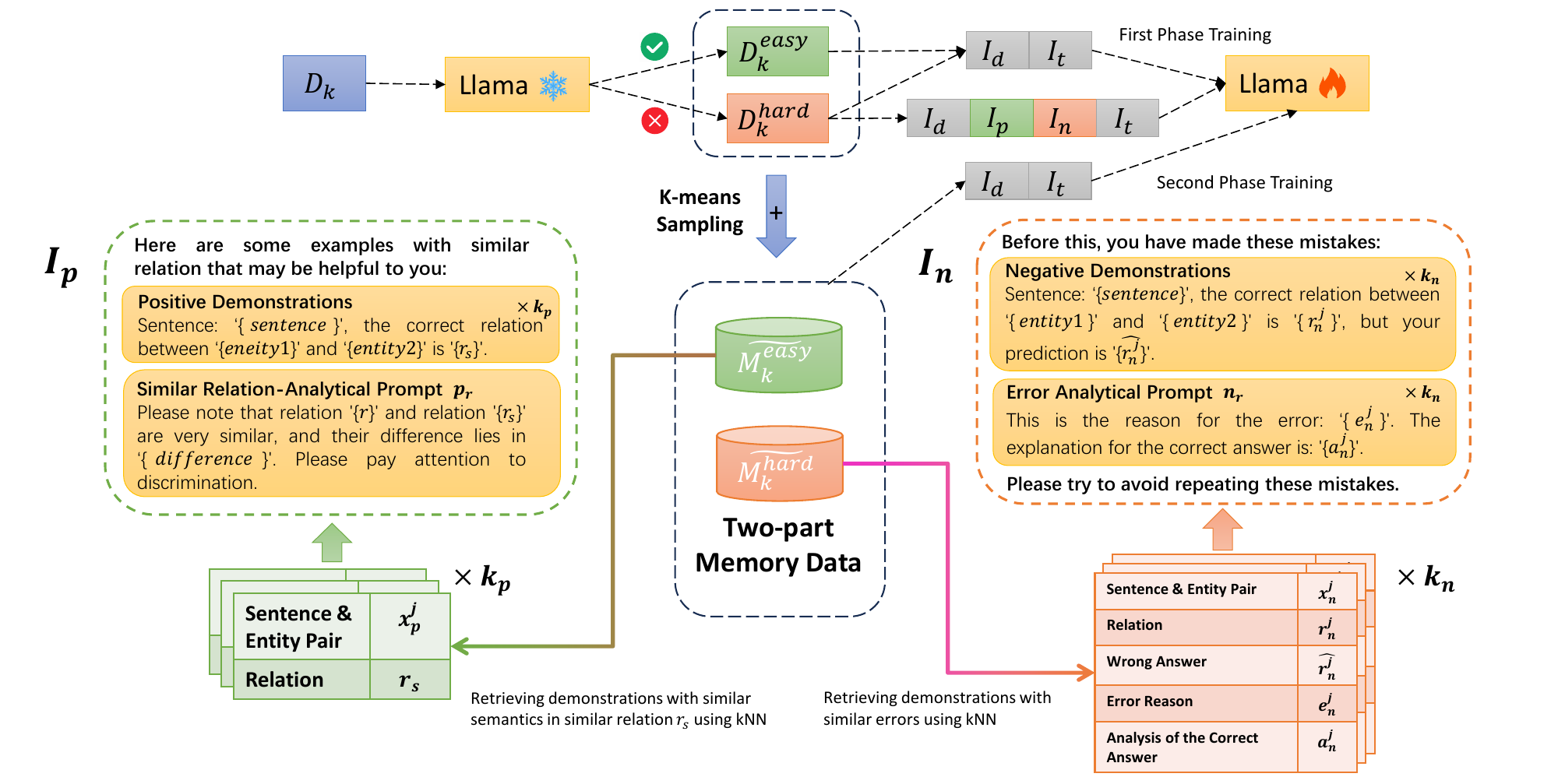}
    \caption{The overall framework of our model. We split the data into easy and hard parts based on the model's inference results on the new task and form the corresponding two-part memory data using the k-means algorithm. For hard samples, we use the contrastive instruction to train the model. In this figure, green represents positive/easy samples, while red represents negative/hard samples. They help the model maintain connections between old and new tasks while correcting cognitive biases from both positive and negative aspects.}
    \label{fig:framework}
\end{figure*}

\subsection{Overall Framework}
The overall framework is shown in Figure \ref{fig:framework}. For a new task $T_k$, we categorize $D_{k}$ into two sets: $D_k^{easy}$ and $D_k^{hard}$, based on whether the model's answers are correct or not. In the initial training phase, we employ a dual-task fine-tuning approach. The first task is to construct simple instructions using all samples to directly train the model. The second task is specific for hard samples, which is to construct contrastive instructions with positive and negative demonstrations to train the model, facilitating it to correct cognitive biases. We then use the k-means algorithm to sample from both simple and hard data to form corresponding memory. In the second training phase, we train the model using all memory to further prevent catastrophic forgetting.

\subsection{Two-part Data Classification}
\subsubsection{Classification of Easy and Hard Samples}
For a new task $T_k$, we first have the model perform inference on the corresponding training data $D_k$. If the predicted answer is correct, the data is categorized as easy samples, indicating knowledge that the model has already mastered. Conversely, if the prediction is incorrect, the data is labeled as hard samples, reflecting knowledge gaps and the model’s current cognitive biases regarding the new task. Following this inference process, we obtain two distinct datasets: $D_k^{easy}$ and $D_k^{hard}$. Then we utilize an external LLM (GPT-3.5-turbo in this paper) to generate an error reason analysis and an analysis of the correct answer for each incorrect prediction, to be used later.

\subsubsection{Two-part Memory Selection}
To help the model recall knowledge from previous tasks and alleviate forgetting issues, we pick a certain amount of data as memory for each relation $r \in R_{k}$. Following previous works \cite{Hu2022ImprovingCR,Zhao2023ImprovingCR_r:12}, we employ the k-means algorithm for sampling. We cluster all instances belonging to relation $r$, with the number of clusters equal to the memory size. Then we select the instances closest to the center of each cluster as the memory samples for that cluster. Unlike previous research, we split the memory data into easy memory data and hard memory data. We sample $m$ instances separately for $D_r^{easy}$ and $D_r^{hard}$, denoted as $M_r^{easy}$ and $M_r^{hard}$, respectively.

\subsection{Contrastive Instruction Tuning}

\subsubsection{Demonstrations Retrieval}
For a hard sample $x^{hard}$ belonging to the relation $r$, in order to construct the contrastive instruction, it is necessary to retrieve positive demonstrations $P$ and negative demonstrations $N$.

For positive demonstrations, we first identify the most similar relation $r_s$ from the previously learned relation set $\tilde{R_{k-1}}$. Following previous works \cite{wang-etal-2022-learning-robust, Xiong2023RationaleEnhancedLM}, we use the average embedding of data for each relation as its representation vector and then calculate the similarity between relations using cosine distance. Then we use the sentence of $x^{hard}$ as the query and retrieve the top $k_p$ demonstrations that are most semantically similar from the easy memory data $M_{r_s}^{easy}$ to form $P=\{(x_p^j, r_s) \}_{j=1}^{k_p}$.

For negative demonstrations, we use the error reason of $x^{hard}$ as the query to retrieve the top $k_n$ demonstrations with the most similar error reasons from the hard memory data $\tilde{M_{k-1}^{hard}}$ that the model has encountered before to form $N=\{(x_n^j, r_n^j,\hat{r_n^j},e_n^j,a_n^j) \}_{j=1}^{k_n}$, where $r_n^j$ represents the correct relation, $\hat{r_n^j}$ represents the relation that the model answered incorrectly, $e_n^j$ represents the error reason in this case, and $a_n^j$ represents the analysis for the correct answer.

\subsubsection{Instruction Construction}
We construct a simple instruction $I_{simple}$ for directly training the model. The instruction consists of two parts: the task description and the prediction instruction. The task description instruction $I_d$ explains the definition of the relation extraction task to the model and prompts it to select from a predefined list of relations.  The format is as follows:

\begin{tcolorbox}[title = {Task Description Instruction}]
\fontsize{10}{12}\selectfont
Now you need to complete the relation extraction task, which is to give you a sentence, two entities in the sentence, and a predefined relation list. You need to tell me what relation exists between these two entities.
\end{tcolorbox}


The prediction instruction $I_t (x,\tilde{R_k})$ informs the model about the data that needs to be predicted and relations in $\tilde{R_k}$.  The format is as follows:

\begin{tcolorbox}[title = {Prediction Instruction}]
\fontsize{10}{12}\selectfont
Given the sentence:“\{sentence\}”, what is the relation between “\{entity1\}” and “\{entity2\}”? Please select from these relations: [\{$\tilde{R_k}$\}].
\end{tcolorbox}


Then we define $I_{simple}$ as
\begin{equation}
    I_{simple} = I_{d}\oplus I_{t}( x,\tilde{R_k} )
\end{equation}
where $\oplus$ represents concatenation.

For $D_r^{hard}$, we enhance the model's ability to avoid errors by further adding positive and negative sample instructions based on $I_{simple}$. For a sample in $D_r^{hard}$, given its retrieved positive demonstrations $P$ and negative demonstrations $N$, we first construct the positive sample instruction $I_p (P,r)$. As shown in Figure \ref{fig:framework}, we inform the model of the sentence, entity pair, and relation of each positive demonstration. In the similar relation-analytical prompt $p_r$, we use an external LLM to generate the differences between relation $r$ and relation $r_s$ of the positive demonstration. This information emphasizes the distinctions between these two similar relations, enabling the model to effectively leverage useful knowledge from the demonstrations while disregarding misaligned information. The generation method for $p_r$ is provided in Appendix \ref{appendix: Analytical Prompt Generation}.

Subsequently, we construct the negative sample instruction $I_n (N)$. As shown in Figure \ref{fig:framework}, we inform the model of each negative demonstration, including the sentence, entity pair, correct relation, and the incorrect relation that the model answered. In addition, the error analytical prompt $n_r$, including the reason for the error and the analysis of the correct answer which are generated by an external LLM, is also provided. The generation method for $n_r$ is provided in Appendix \ref{appendix: Analytical Prompt Generation}. This prompt helps the model understand the errors more comprehensively and avoid repeating similar mistakes.

We define $I_{contrastive}$ as
\begin{equation}
    I_{contrastive} = I_{d}\oplus I_p (P,r)\oplus I_n (N)\oplus I_{t}( x,\tilde{R_k} ),
\end{equation}
where $\oplus$ represents concatenation.

Some instruction examples are provided in Appendix \ref{appendix: Instruction Examples}.

\subsubsection{Dual-task Fine-tuning}
For a new task $T_k$, after completing the division of data into easy and hard samples and retrieving positive and negative samples from $D_k$, we train the model through instruction tuning.

In the initial training phase, we construct two instruction tuning tasks, \(I_{simple} \rightarrow r\) and \(I_{contrastive} \rightarrow r\). The first task directly trains the model using both easy and hard samples, while the second task employs hard samples along with contrastive instructions to enhance the model's ability to leverage learned knowledge and avoid errors. The motivation behind this design is that previous research \cite{Xiong2023RationaleEnhancedLM, Zelikman2024QuietSTaRLM} has shown that multi-task fine-tuning allows models to grasp the real rationale, thereby enhancing their performance.

\subsection{Memory Replay} In the second training phase, we fine-tune the model using all the memory data $\hat{M_k}$ with $I_{simple}$, further mitigating the issue of catastrophic forgetting. Finally, we use $I_{simple}$ to have the model make predictions on the test set of $T_k$.

\section{Experiments}

\begin{table*}[t]  
\centering
\resizebox{\linewidth}{!}{
\begin{tabular}{l c c c c c c c c c c}
& & & & & & & & & &\\
\toprule
\multicolumn{11}{c}{\textbf{TACRED}}\\
\midrule
\textbf{Models} & \textbf{T1} & \textbf{T2} & \textbf{T3} & \textbf{T4} & \textbf{T5} & \textbf{T6} & \textbf{T7} & \textbf{T8} & \textbf{T9} & \textbf{T10}\\
\midrule
RP-CRE  & 97.6 & 90.6 & 86.1 & 82.4 & 79.8 & 77.2 & 75.1 & 73.7 & 72.4 & 72.4\\
ACA  & 98.0 & 92.1 & 90.6 & 85.5 & 84.4 & 82.2 & 80.0 & 78.6 & 78.8 & 78.1 \\
CRL  & 97.7 & 93.2 & 89.8 & 84.7 & 84.1 & 81.3 & 80.2 & 79.1 & 79.0 & 78.0\\
CDec  & 97.7 &	92.8 &	91.0 &	86.7 &	85.2 &	82.9 &	80.8 &	80.2 &	78.8 &	78.6 \\
CEAR  & 97.7 &	94.3 &	92.3 &	88.4 &	86.6 &	84.5 &	82.2 &	81.1 &	80.1 &	79.1\\
RationaleCL  & 98.6 &	94.4 &	91.5 &	88.1 &	86.5 &	84.9 &	84.5 &	82.5 &	81.6 &	80.8\\
EoE  & \textbf{98.7} &	94.7 &	90.6 &	87.8 &	87.2 &	85.9 &	84.3 &	83.2 &	82.7 &	81.5\\
DP-CRE  & 97.8 &	93.8 &	91.5 &	87.5 &	85.7 &	84.2 &	82.9 &	81.3 &	81.5 &	80.7\\
\midrule
CIT-CRE & 98.6\footnotesize{±0.6} &	\textbf{95.0}\footnotesize{±2.1} &	\textbf{92.7}\footnotesize{±3.1} &	\textbf{89.1}\footnotesize{±2.3} &	\textbf{88.1}\footnotesize{±1.8} &	\textbf{87.0}\footnotesize{±2.6} &	\textbf{85.0}\footnotesize{±1.2} &	\textbf{83.9}\footnotesize{±1.6} &	\textbf{83.8}\footnotesize{±1.1} &	\textbf{82.6}\footnotesize{±0.8}\\
\bottomrule
\toprule
\multicolumn{11}{c}{\textbf{FewRel}}\\
\midrule
\textbf{Models} & \textbf{T1} & \textbf{T2} & \textbf{T3} & \textbf{T4} & \textbf{T5} & \textbf{T6} & \textbf{T7} & \textbf{T8} & \textbf{T9} & \textbf{T10}\\
\midrule
RP-CRE  & 97.9 & 92.7 & 91.6 & 89.2 & 88.4 & 86.8 & 85.1 & 84.1 & 82.2 & 81.5\\
ACA  & 98.3 & 95.0 & 92.6 & 91.3 & 90.4 & 89.2 & 87.6 & 87.0 & 86.3 & 84.7 \\
CRL  & 98.1 & 94.6 & 92.5 & 90.5 & 89.4 & 87.9 & 86.9 & 85.6 & 84.5 & 83.1\\
CDec  & 98.4 & 95.4 & 93.2 & 92.1 & 91.0 & 89.7 & 88.3 & 87.4 & 86.4 & 84.8\\
CEAR  & 98.1 & \textbf{95.8} & 93.6 & 91.9 & 91.1 & 89.4 & 88.1 & 86.9 & 85.6 & 84.2\\
RationaleCL & 98.6 & 95.7 & 93.4 & 92.3 & 91.3 & 89.7 & 88.2 & 87.3 & 86.3 & 85.1\\
EoE  & 97.8 & 95.0 & 93.6 & 92.5 & 91.6 & 90.0 & 88.9 & 87.9 & 86.9 & 85.5\\
DP-CRE  & 98.5 & 95.4 & \textbf{93.7} & 92.1 & 90.9 & 89.4 & 88.5 & 87.4 & 86.3 & 85.1\\
\midrule
CIT-CRE & \textbf{99.1}\footnotesize{±0.5} & 95.3\footnotesize{±1.7} & 93.5\footnotesize{±1.8} & \textbf{93.2}\footnotesize{±0.9} & \textbf{92.7}\footnotesize{±2.1} & \textbf{91.2}\footnotesize{±1.5} & \textbf{90.8}\footnotesize{±1.9} & \textbf{90.2}\footnotesize{±1.3} & \textbf{89.3}\footnotesize{±1.4} & \textbf{89.0}\footnotesize{±0.5} \\
\bottomrule
\end{tabular}
}
\caption{Accuracy (\%) on all seen relations after learning each task. The result of RationaleCL, CEAR, CRL, ACA and RP-CRE are obtained from the RationaleCL's original paper, and the results of other baselines are directly cited from their original papers.
We conducted experiments using the same task sequence as previous research. Each task sequence consists of 10 subsets of relation types, and the average results of five different task sequences are reported for each experiment.
We report the average accuracy and standard deviations of 5 different experiments.  
We show the best results in \textbf{boldface}.}
\label{table:1}
\end{table*}

\subsection{Experiment Setup}

\paragraph{Datasets.} We conduct our experiments on two mainstream benchmark datasets that are widely used in previous CRE works. (1) \textbf{FewRel} \cite{Han2018FewRelAL} is a benchmark dataset initially proposed for few-shot RE. It contains 100 relations, each with 700 samples. To be consistent with previous CRE work \cite{Xiong2023RationaleEnhancedLM}, we use its version containing 80 relations. (2) \textbf{TACRED} \cite{TACRED} is a large RE benchmark dataset that includes 42 relations (including \textit{no\_relation}) and 106,264 samples collected from news and web documents. To be consistent with previous work \cite{Xiong2023RationaleEnhancedLM}, we remove \textit{no\_relation} in our experiments. To address the issue of sample imbalance, we limited the number of training samples per relation to 320 and the number of test samples to 40, which is also consistent with previous work.

\paragraph{Compared Models.} We compare our model with the following baselines: \textbf{RP-CRE} \cite{Cui2021RefiningSE_r:25}, which refines sample embeddings using relation prototypes and memory networks; \textbf{ACA} \cite{wang-etal-2022-learning-robust}, which employs adversarial class augmentation for robust representations; \textbf{CRL} \cite{Zhao2022ConsistentRL}, which uses contrastive learning and knowledge distillation for stable relation embeddings; \textbf{CDec} \cite{Xia2023EnhancingCR}, a classifier decomposition framework for robust representation learning; \textbf{CEAR} \cite{Zhao2023ImprovingCR_r:12}, which addresses overfitting with memory-insensitive prototypes and augmentation; \textbf{RationaleCL} \cite{Xiong2023RationaleEnhancedLM}, which leverages multi-task rationale tuning and contrastive replay; \textbf{EoE} \cite{zhou-etal-2024-ensemble}, a rehearsal-free framework; and \textbf{DP-CRE} \cite{huang-etal-2024-dp}, which decouples prior knowledge preservation and new knowledge acquisition.

\paragraph{Experimental Settings.} We use Llama3-8B-Instruct \cite{llama3modelcard} as the backbone and fine-tune it using the parameter-efficient tuning method LoRA \cite{Hu2021LoRALA}. We use AdamW as the optimizer. The batch size for training is set to 32, with 2 epochs for each of the two stages, and the learning rate is set to 3e-5. We set the size of both the easy and hard memory to 5, so the total size is 10, which is consistent with other baselines except for EoE which does not require memory. We set $k_p$ and $k_n$ to 3. Our experiments are conducted on a single NVIDIA A100 GPU.

\subsection{Main Results} 
The performances of our model and baselines are shown in Table \ref{table:1}. Notably, since we use LoRA, the number of trainable parameters is approximately 21 million, so the comparison is fair. Based on the results, the following observations can be drawn:

Our method enables the model to self-correct cognitive biases by exploiting error cases, achieving SOTA performance on both CRE datasets. This suggests that our method effectively mitigates catastrophic forgetting.

As the model learns more tasks, our method shows increasingly significant performance advantages. In TACRED’s T1, FewRel’s T2 and T3, our method performs slightly worse compared to other baselines. This is attributed to the limited diversity of samples in the memory at early stages, resulting in insufficiently rich retrieval data. Furthermore, the model has not fully mastered learning from error cases, leading to suboptimal performance. Nevertheless, as the task sequence progresses, our method demonstrates enhanced adaptability, outperforming the best baselines by 1.1\% and 3.5\% in TACRED and FewRel’s T10, respectively. This is because we not only utilize memory for replay but also to establish connections between new and old tasks through contrastive instructions. As the memory accumulates, the model gains access to a more diverse dataset, resulting in a gradual slowdown in performance decline.

In addition, all models perform relatively poorly on TACRED, and our model also shows a higher standard deviation on this dataset. This is because TACRED is class-imbalanced and contains fewer training samples, making it more difficult. In contrast, FewRel covers a broader range of relation types and more data, enabling our model to collect more diverse memory sets and utilize them. As a result, our model achieves more outstanding performance on FewRel compared to other baselines.


\subsection{Ablation Study}

\begin{table}[htbp]
\centering
\resizebox{\linewidth}{!}{
\begin{tabular}{l c c c c c}
\toprule
\textbf{Models} & T6 & T7 & T8 & T9 & T10 \\
\midrule
Llama3-8b & \textbf{87.0} & \textbf{85.0} & \textbf{83.9} & \textbf{83.8} & \textbf{82.6} \\
Qwen2.5-7b & 86.5 &	83.9 &	82.8 &	81.9 &	81.4 \\
Qwen2.5-3b & 85.6 &	82.9 &	82.1 &	81.3 &	80.7 \\
\bottomrule
\end{tabular}
}
\caption{Results of different PrLMs on TACRED.}
\label{tab:dif}
\end{table}

\begin{table}[htbp]
\centering
\resizebox{\linewidth}{!}{
\begin{tabular}{l c c c c c}
\toprule
\textbf{Models} & T6 & T7 & T8 & T9 & T10 \\
\midrule
CIT-CRE & \textbf{87.0} & \textbf{85.0} & \textbf{83.9} & \textbf{83.8} & \textbf{82.6} \\
\midrule
w/o $D_k^{hard}$ & 86.2 &	83.9 &	82.7 &	82.2 &	81.7 \\
\midrule
w/o $I_p$ & 86.3 &	84.5 &	83.3 &	83.2 &	82.5 \\
w/o $I_n$ & 86.2 &	84.2 &	83.5 &	83.1 &	82.2 \\
w/o $I_p+I_n$ & 86.3 &	84.5 &	83.3 &	83.2 &	82.5 \\
\midrule
w/o $p_{r}$ & 86.3 &	84.0 &	83.1 &	82.6 &	81.9 \\
w/o $n_{r}$ & 86.8 &	84.6 &	83.8 &	82.9 &	82.4 \\

\bottomrule
\end{tabular}
}
\caption{Ablation study results on TACRED.}
\label{tab:abl}
\end{table}

\subsubsection*{Different PrLMs}

We employ different pre-trained models (Qwen2 series \cite{qwen2}) as our backbone for training and testing on TACRED. Table \ref{tab:dif} shows that the performance of other models declines to varying degrees as the number of parameters decreases, yet they remain competitive. This proves the effectiveness of exploiting error cases.

\subsubsection*{Exploiting Error Cases Brings Improvement}
To validate the effectiveness of leveraging error cases, we conduct an ablation experiment. As shown in Table \ref{tab:abl}, for ``w/o $D_k^{hard}$", we no longer distinguish between hard and easy cases, that is, we only use $I_{simple}$ to train the model on all data. The results reveal a noticeable performance degradation, which proves the effectiveness and importance of focusing on error cases in CRE tasks.

\subsubsection*{The Importance of the Contrastive Instruction}

To validate the effectiveness of the contrastive instruction, we design ablation experiments by systematically removing relevant components. Specifically, we examine the effects of removing $I_p$, $I_n$, and both simultaneously on model performance. The results, as shown in Table \ref{tab:abl}, demonstrate that compared to removing both demonstrations simultaneously, retaining a certain part alone does not improve the model's performance. We attribute this to the lack of contrast. Negative demonstrations are selected from hard memory data, which contains more complex knowledge. Therefore, when the model only has access to negative demonstrations, it can learn to avoid mistakes but may not fully understand how to correctly answer the current error case. Conversely, positive demonstrations are selected from easy memory data and contain simpler knowledge. While the model can transfer knowledge from these cases to the current task, it may struggle to understand the underlying causes of errors, leading to repeated mistakes. This highlights the importance of the contrast between positive and negative demonstrations, as it helps the model to better correct cognitive biases. 

\subsubsection*{Analytical Prompt Help Transfer Knowledge}
In addition, we also investigate the effectiveness of the analytical prompt $p_r$ and $n_r$, as shown in Table \ref{tab:abl}. We find that removing them can also cause a significant performance decrease. We believe this is because inference information helps the model understand how to transfer the knowledge in the demonstration to the current problem. Additionally, we find that removing $p_r$ leads to a more significant performance drop compared to removing $n_r$. We consider that this is because, without $p_r$, the model is more likely to be misled by demonstrations from similar relations and fails to effectively utilize the knowledge within them.

\section{Analysis}

\subsection{The Influence of Memory Size}

\begin{figure}[htbp]
    \centering
    \includegraphics[width=1\linewidth]{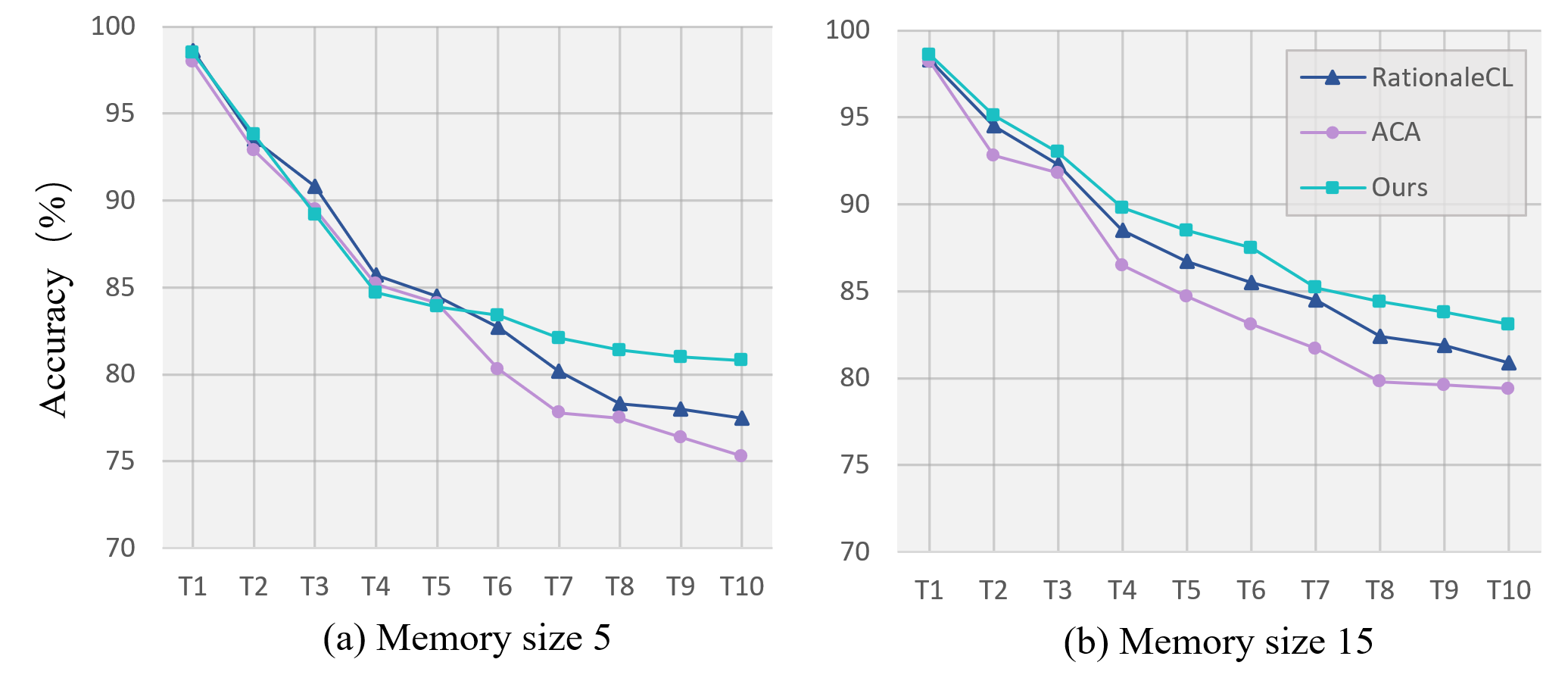}
    \caption{Comparison of model's performance on different memory sizes on TACRED.}
    \label{figure:mem}
\end{figure}

Memory size is defined as the typical number of samples stored for each relation. For memory-based CRE models, memory size can significantly affect model performance. We reduce and increase the memory size separately based on the main experiment, and figure \ref{figure:mem} shows the performance of our model compared to RationaleCL and ACA at different memory sizes. It can be observed that the performance of the CRE model is positively correlated with memory size, highlighting the importance of memory size. Additionally, as the number of tasks increases, our model demonstrates increasingly significant advantages, which become even more pronounced when the memory size is 5. We attribute this to our method's more efficient utilization of limited memory data. Our approach enables the model to understand the relationships between the current task and previous tasks, rather than merely replaying the data. These experiments demonstrate that our model is robust to changes in memory size.

\subsection{Are More Demonstrations Better?}

\begin{table}[htbp]
\centering
\begin{tabular}{l c c c c c}
\toprule
\textbf{Number} & T6 & T7 & T8 & T9 & T10 \\
\midrule
1 & 85.9 &	84.5 &	83.5 &	83.1 &	81.9 \\
3 & \textbf{87.0} & \textbf{85.0} & \textbf{83.9} & \textbf{83.6} & \textbf{82.6} \\
5 & 86.6 &	84.7 &	83.5 &	\textbf{83.6} &	82.1 \\
\bottomrule
\end{tabular}
\caption{Results of different demonstration numbers on TACRED.}
\label{tab:number}
\end{table}

The number of demonstrations is crucial for correcting cognitive biases. We set the number of positive and negative demonstrations to 1, 3, and 5 to investigate their impact. The results presented in Table \ref{tab:number} indicate that the optimal performance is achieved with 3 demonstrations. We believe that with only 1 demonstration, the quantity is insufficient, preventing the model from fully understanding how to correct the bias. Conversely, when the number is set to 5, the model's limited memory capacity leads to the inclusion of some irrelevant data in the demonstrations, hindering the model's ability to extract useful information and resulting in decreased performance.

\subsection{Case Study}

\begin{figure}
    \centering
    \includegraphics[width=1\linewidth]{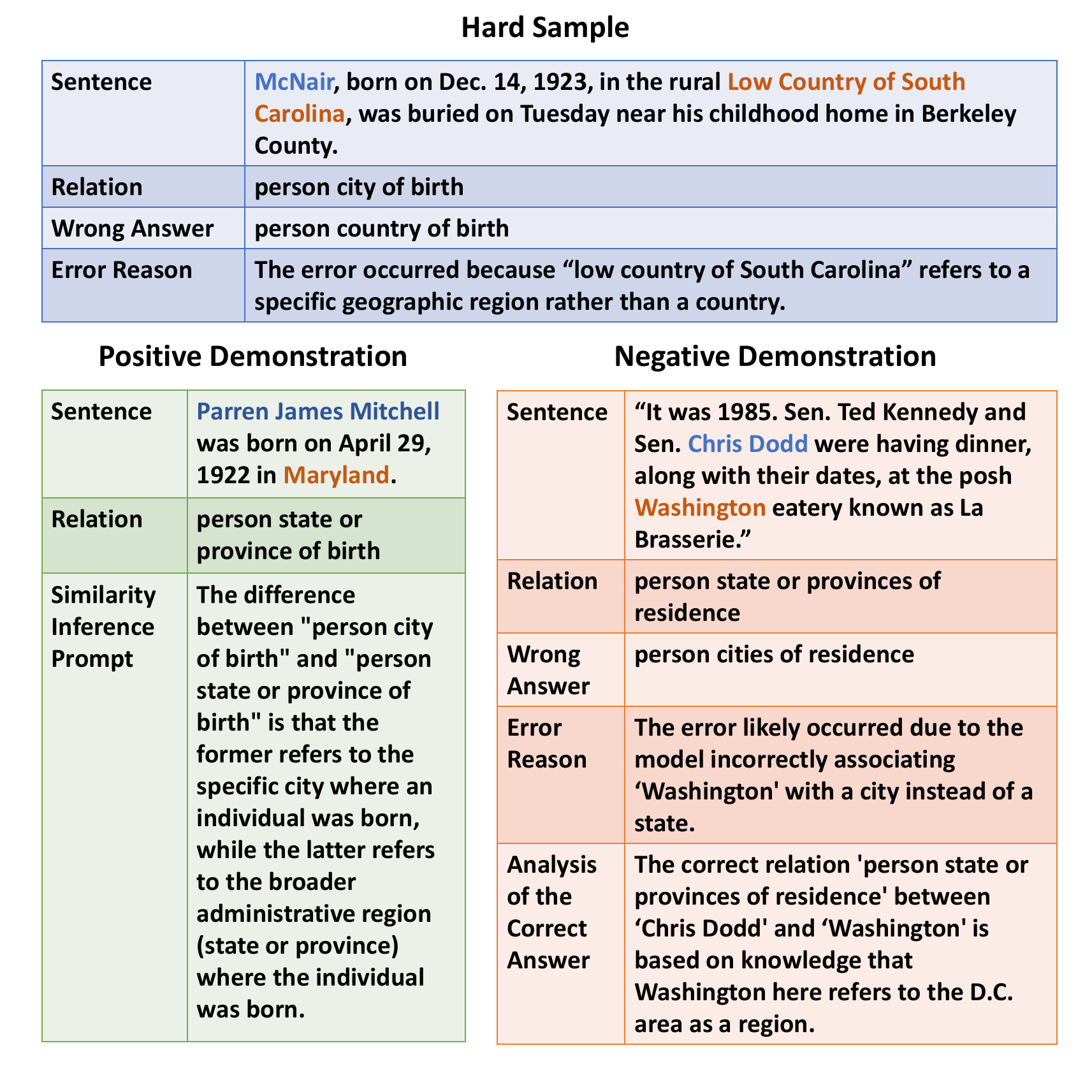}
    \caption{A case study on TACRED.}
    \label{fig:case}
\end{figure}

To more intuitively demonstrate how our method works, we present a case in figure \ref{fig:case}. For a hard sample, the model did not correctly identify the ``person city of birth" relation. To correct the cognitive bias, the model retrieves a semantically similar positive demonstration from the easy memory data of the similar relation ``person state or province of birth" and guides the model to transfer knowledge about this data through a similar relation-analytical prompt. Additionally, the model retrieves a negative demonstration from the hard memory data based on the similarity of the error reasons. The negative demonstration provides a similar error case and explains the error reason and the correct answer, guiding the model to avoid these mistakes. By contrasting these demonstrations, the model can better correct the current cognitive bias and finally make the correct prediction.

\section{Conclusion}
In this paper, we propose a method for CRE based on LLM. Our approach focuses on the error samples generated by the model during the learning process and utilizes these samples by dividing the training data and memory data into two parts for differentiation processing. In addition, by devising a continual contrastive instruction tuning strategy, our approach enables the LLM to continually leverage previous knowledge to correct current cognitive biases by utilizing easy and hard memory data. The contrastive instruction establishes a connection between new and old data, thus improving the model's generalization and transfer capabilities. Experiments on TACRED and FewRel show that our method outperforms previous state-of-the-art models with significant improvements, demonstrating the importance of specializing in exploiting error cases. In the future, we will continue to explore methods that exploit error cases based on LLM in other continual learning tasks.

\section*{Limitations}
We consider that our paper has several limitations: 1) Our method mainly uses instruction-tuning to develop error cases, which requires the backbone model to have instruction understanding ability, which may limit the generalization of our approach. In addition, this also requires relatively high computing power. 2) Our experiment utilizes an external LLM to analyze the error causes, which increases the dependency on calling the API.

\bibliography{anthology,custom}
\bibliographystyle{acl_natbib}

\appendix


\section{Analytical Prompt Generation}
\label{appendix: Analytical Prompt Generation} 
We use an external LLM (GPT-3.5-turbo in this paper) to generate the similar relation-analytical prompt $p_{r}$ for positive demonstrations, and the error analytical prompt $n_{r}$ for negative demonstrations.
For $p_{r}$, we randomly select two samples from the easy memory data corresponding to two relations to display to the model and prompt it to analyze the differences between the two relations. The prompt is as follows:

\begin{tcolorbox}[title = {$p_{r}$ Generation Prompt}]
\fontsize{10}{12}\selectfont 
I am doing a relation extraction task. Now I have two very similar relations for you. Please tell me the difference between them in one sentence. For the first sentence: `\{sentence\_1\}', the relation between `\{entity1\_1\}' and `\{entity2\_1\}' is `\{relation\_1\}'; For the second sentence: `\{sentence\_2\}', the relation between `\{entity1\_2\}' and `\{entity2\_2\}' is `\{relation\_2\}'; Now please tell me the difference between `\{relation\_1\}' and `\{relation\_2\}' in one sentence.
\end{tcolorbox}


For $n_{r}$, we inform the model of the sentence, entity pairs, the correct relation, and the incorrect relation previously provided by the model. We then prompt it to analyze the reason for the error and provide an analysis for the correct answer. The prompt is as follows:

\begin{tcolorbox}[title = {$n_{r}$ Generation Prompt}]
\fontsize{10}{12}\selectfont 
For this sentence: `\{sentence\}', the correct relation between `\{entity1\}' and `\{entity2\}' should be `\{correct\_relation\}', but it is predicted to be `\{wrong\_relation\}'. Please analyze the reason for the error in one sentence and provide an analysis of the correct answer in one sentence.
\end{tcolorbox}

Examples are provided in Appendix \ref{appendix: Instruction Examples}.



\section{Instruction Examples}
\label{appendix: Instruction Examples} 

In this section, we provide some examples of instructions. Table \ref{table:5} shows an example of $I_{simple}$ constructed using an easy sample from TACRED. Table \ref{table:6} shows an example of $I_{contrastive}$ using a hard sample from TACRED. Table \ref{table:7} and Table \ref{table:8} respectively show the generation process of $p_r$ and $n_r$, both using data from TACRED as a demonstration.


\begin{table*}[t]  
\centering
\begin{tabularx}{\textwidth}{X} 
\toprule
\textbf{Task Description Instruction} \\
Now you need to complete the relation extraction task, which is to give you a sentence, two entities in the sentence, and some predefined relations lists. You need to tell me what relation exists between these two entities. \\
\midrule
\textbf{Positive Demonstrations} \\
Here are some examples with similar relations that may be helpful to you: [sentence: “Parren James Mitchell was born on April 29, 1922, in Maryland.”, the correct relation between “Parren James Mitchell” and “Maryland” is “person state or province of birth”.sentence: “Ahearn was born Oct. 7, 1954, in Nashville, Tenn., and graduated with honors from the University of Alabama.”, the correct relation between “Ahearn” and “Tenn.” is “person state or province of birth”. sentence: “Born in 1955 in Montgomery, Alabama, King was just an infant when her home was bombed during the turbulent civil rights era.”, the correct relation between “her” and “Alabama” is “person state or province of birth”.]. Please note that the relation “person state or province of birth” and the relation “person city of birth” are very similar, and their difference lies in “The ‘person city of birth’ relation specifies the specific city where a person was born, while the ‘person state or province of birth’ relation indicates the broader state or province where a person was born.”. Please pay attention to discrimination. \\
\midrule
\textbf{Negative Demonstrations} \\
Before this, you have made these mistakes: [sentence: “it was 1985 ... sen. ted kennedy and sen. chris dodd were having dinner, along with their dates at the posh washington eatery known as la brasserie.”, the correct relation between “chris dodd” and “washington” is “person state or provinces of residence”, but your prediction is “person cities of residence”. This is the reason for the error: “The error likely occurred due to the model incorrectly associating ‘washington’ with a city instead of a state.”. The reason for the correct answer is: “The correct relation ‘person state or provinces of residence’ between ‘chris dodd’ and ‘washington’ is based on knowledge that Washington here refers to the D.C. area as a region.”.sentence: “an arms control expert and political science professor at city college of new york, forsberg launched the movement in 1980 when she wrote the `` call to halt the nuclear arms race , '' a position paper that outlined the devastating potential of the arsenals possessed by the united states and what was then the soviet union .”, the correct relation between “forsberg” and “new york” is “person state or provinces of residence”, but your prediction is “person cities of residence”. This is the reason for the error: “The error may be due to the model incorrectly associating ‘new york’ with ‘city’, confusing it with a city rather than a state or province.”. The reason for the correct answer is: “‘new york’ should be correctly associated as the state where ‘city college of new york’ is located, indicating the person's state of residence.”. sentence: “deaver formed his own company after reagan left the state capital -- the former governor and presidential aspirant was his chief client -- and then joined reagan in washington after his 1980 election .”, the correct relation between “his” and “washington” is “person state or provinces of residence”, but your prediction is “person cities of residence”. This is the reason for the error: “The error likely occurred due to the model misinterpreting ‘washington’ as a city instead of the state of Washington.”. The reason for the correct answer is: “The correct relation is ‘person state or provinces of residence’, where ‘his’ refers to Washington, the state, not the city.”.], please try to avoid repeating these mistakes. \\
\midrule
\textbf{Prediction Instruction} \\
Now given the sentence: “McNair, born on Dec. 14, 1923, in the rural low country of South Carolina, was buried on Tuesday near his childhood home in Berkeley county.”, what is the relation between “McNair” and “low country of south carolina”? Please select from these relations: [organization members, person cities of residence, ..., person state or province of death, person siblings], and strictly return the answer in the following JSON format:\{“relation”: xxx\}. \\

\bottomrule
\end{tabularx}
\caption{An example of a hard case.}
\label{table:5}
\end{table*}

\begin{table*}[t]  
\centering
\begin{tabularx}{\textwidth}{X} 
\toprule
\textbf{Task Description Instruction} \\
Now you need to complete the relation extraction task, which is to give you a sentence, two entities in the sentence, and some predefined relations lists. You need to tell me what relation exists between these two entities. \\
\midrule
\textbf{Prediction Instruction} \\
Now given the sentence: “the 33-year-old Carson says he doesn't believe his religious identity hurts him politically.”, what is the relation between “Carson” and “33-year-old”? Please select from these relations: [organization alternate names, organization country of headquarters, organization number of employees members, person state or province of birth, person date of birth, person age, organization members, person parents, organization website, person origin, person city of death, person title, person city of birth, person schools attended, person cities of residence, person charges, person country of birth, organization dissolved, person alternate names, organization top members employees, person state or provinces of residence, person employee of, organization founded by, organization member of, organization subsidiaries, organization founded, person siblings, person date of death, person countries of residence, person state or province of death, person spouse, organization state or province of headquarters, person other family, person children, person cause of death, organization political religious affiliation], and strictly return the answer in the following JSON format:\{“relation”:xxx\}. \\
\bottomrule
\end{tabularx}
\caption{An example of an easy case.}
\label{table:6}
\end{table*}

\begin{table*}[t]  
\centering
\begin{tabularx}{\textwidth}{X} 
\toprule
\textbf{$p_{r}$ Generation Prompt} \\
I am doing a relation extraction task. Now I have two very similar relations for you. Please tell me the difference between them in one sentence. For the first sentence: “McNair, born on Dec. 14, 1923, in the rural low country of South Carolina, was buried on Tuesday near his childhood home in Berkeley county.”, the relation between “McNair” and “low country of South Carolina” is `person city of birth'; For the second sentence: “Parren james mitchell was born on April 29, 1922, in Maryland.”, the relation between “Parren james mitchell” and “Maryland” is “person state or province of birth”; Now please tell me the difference between “person city of birth” and “person state or province of birth” in one sentence. Please return the answer in JSON format as follows: \{“difference”: xxx\}.\\
\midrule
\textbf{GPT Response} \\
\{“difference”: The ‘person city of birth’ relation specifies the specific city where a person was born, while the ‘person state or province of birth’ relation indicates the broader state or province where a person was born.\} \\
\bottomrule
\end{tabularx}
\caption{An example of the $p_{r}$ Generation Prompt.}
\label{table:7}
\end{table*}

\begin{table*}[t]  
\centering
\begin{tabularx}{\textwidth}{X} 
\toprule
\textbf{$n_{r}$ Generation Prompt} \\
For this sentence: “McNair, born on Dec. 14, 1923, in the rural low country of South Carolina, was buried on Tuesday near his childhood home in Berkeley county.”, the correct relation between `McNair' and `low country of South Carolina' should be ‘person city of birth’, but it is predicted to be ‘person country of birth’. Please analyze the reason for the error in one sentence and provide an analysis of the correct answer in one sentence. Please return the answer in JSON format as follows:\{“error\_reason”: xxx, “correct\_answer\_analysis”: xxx\}.\\
\midrule
\textbf{GPT Response} \\
\{“error\_reason”: The error is likely due to the model incorrectly associating “low country of South Carolina” as a country instead of a city due to lack of context or training data.\\
“correct\_answer\_analysis”: ‘McNair’ and ‘low country of South Carolina’ should have a ‘person city of birth’ relation because the ‘low country of South Carolina’ refers to a specific region within South Carolina where the person was born.\} \\
\bottomrule
\end{tabularx}
\caption{An example of the $n_{r}$ Generation Prompt.}
\label{table:8}
\end{table*}

\end{document}